%% file: acl_latex.tex
\documentclass[11pt]{article}

\usepackage[final]{acl}

\usepackage{times}
\usepackage{latexsym}

\usepackage[T1]{fontenc}

\usepackage[utf8]{inputenc}

\usepackage{microtype}

\usepackage{inconsolata}

\usepackage{graphicx}

\usepackage{booktabs}
\usepackage{multirow} 

\title{Parallel Thinking, Sequential Answering: Bridging NAR and AR for Efficient Reasoning}

\author{Qihang Ai \\
  Nanyang Technological University \\
  \texttt{qihang005@e.ntu.edu.sg} \\\And
  Haiyun Jiang\thanks{Corresponding author.} \\
  Shanghai Jiao Tong University \\
  \texttt{haiyun2025@sjtu.edu.cn} \\}

\begin{document}
\maketitle
\begin{abstract}
We study reasoning tasks through a framework that integrates auto-regressive (AR) and non-autoregressive (NAR) language models. AR models, which generate text sequentially, excel at producing coherent outputs but often suffer from slow inference—particularly in reasoning-intensive domains such as mathematics and code, where lengthy chains of thought are required. In contrast, NAR models, such as discrete diffusion models, allow parallel generation and offer substantial speedups, though typically at the cost of reduced output quality. To address these limitations, we introduce a new paradigm in which an NAR model efficiently produces intermediate reasoning traces, which subsequently guide an AR model to deliver precise final answers. Experiments demonstrate that our approach yields significant 26\% improvements over strong baselines while substantially reducing inference cost. 
\end{abstract}

\input{latex/intro}
\input{latex/related_work}
\input{latex/experiment}

\bibliography{custom}

\appendix

\input{latex/appendix}

\end{document}

%% file: latex/intro.tex
\section{Introduction}

In recent years, Large Language Models (LLMs) with an emphasis on reasoning—often termed Large Reasoning Models (LRMs)~\cite{xu2025towards}—have achieved significant progress on complex reasoning tasks. A representative example is DeepSeek-R1~\cite{guo2025deepseek}, which employs reinforcement learning (RL) with outcome-based and verifiable rewards to explicitly encourage the development of reasoning skills.

DeepSeek-R1 and its distilled variants generally adopt a structured generation scheme following a \texttt{<think>}–\texttt{<answer>} format. In this framework, the \texttt{<think>} stage produces detailed step-by-step reasoning traces, which then guide the \texttt{<answer>} stage to yield the final solution. This procedure aligns with the widely used Chain-of-Thought (CoT) prompting paradigm~\cite{wei2022chain}, and models adopting such an approach are commonly referred to as R1-style models.

Building on this observation, an increasingly popular inference paradigm has been to enhance model performance by allocating more tokens to the reasoning stage~\cite{luo2025deepscaler,wu2024inference}. While this strategy can improve solution quality, it also introduces practical challenges: the generation of excessively long reasoning sequences increases computational overhead and inference latency. This overproduction of reasoning tokens often leads to the so-called overthinking problem, where many intermediate steps turn out to be redundant or uninformative~\cite{sui2025stop,kumar2025overthink}.

To address the inefficiency stemming from prolonged reasoning, recent research has shifted toward more economical reasoning strategies, including hybrid methods~\cite{anthropic2025claude37,yang2025qwen3} that balance explicit reasoning with concise inference~\cite{renze2024benefits,xu2025chain}, thereby mitigating the high computational cost associated with extended token generation.

However, in challenging reasoning tasks—such as mathematics and programming—that inherently demand multi-step reasoning, explicit thinking becomes indispensable. The Chain-of-Thought paradigm has been widely adopted to enhance the ability of LLMs in tackling these problems~\cite{xu2025chain,kojima2022large}. In addition, some works~\cite{shao2024deepseekmath,gao2023pal} highlight a complementary line of research on code training benefits for mathematical reasoning.

Non-autoregressive (NAR) generation offers an alternative to traditional left-to-right decoding by allowing tokens to be predicted and refined in parallel. A key line of work in this paradigm is diffusion language models (DLMs)~\cite{nie2025large,gongscaling,ye2024diffusion,shao2024deepseekmath,song2025seed,geminiDiffusion2025,yang2025mmada}, which cast text generation as a denoising process that iteratively transforms noise into fluent outputs. Such models bring several benefits for reasoning: parallel generation substantially reduces inference latency, iterative refinement provides a natural mechanism to correct low-confidence predictions, and the bidirectional conditioning strengthens global coherence.
Recent progress, such as Seed Diffusion~\cite{song2025seed}, Mercury Coder~\cite{mercuryCoder2025}, and Gemini Diffusion~\cite{geminiDiffusion2025}, has further demonstrated that large-scale DLMs can achieve both high-quality outputs and high-speed inference.

Inspired by these advantages, we propose a hybrid reasoning paradigm that divides labor between NAR and AR models. Specifically, the NAR component generates the \texttt{think} stage—compact but explicit reasoning traces—while the AR component produces the \texttt{answer} stage, ensuring precise and faithful final outputs. This design inherits the complementary strengths of both approaches: the efficiency and global context modeling of NAR, together with the reliability and expressiveness of AR. In summary, our contribution is a new reasoning framework that combines NAR and AR models to achieve higher accuracy while significantly reducing inference cost.





%% file: latex/related_work.tex
\section{Related Work}

\subsection{Diffusion Language Models}
Autoregressive (AR) models dominate current large language models such as ChatGPT~\cite{gpt4}, generating text in a strictly sequential token-by-token manner. 
However, this token-by-token paradigm inherently limits inference speed. 
Diffusion Language Models (DLMs) address this limitation by enabling efficient parallel generation. 
Through iterative denoising, DLMs can produce multiple tokens—or even full sequences—simultaneously, offering higher inference throughput and better alignment with modern parallel computing hardware~\cite{li2025survey}.

\subsection{Mathematical Reasoning}
Recent efforts have significantly advanced LLMs’ performance on mathematical reasoning benchmarks. Two major directions have emerged: Chain-of-Thought prompting for enhancing multi-step reasoning~\cite{wei2022chain,kojima2022large}, and code-based approaches that leverage program execution to address LLMs’ computational limitations~\cite{gao2023pal,zhou2023solving}. Building on these insights, we emphasize the complementary roles of explicit reasoning and executable code~\cite{roziere2023code} in solving math problems, while also acknowledging math-oriented pre-training as a means of improving general reasoning ability.

%% file: latex/experiment.tex
\section{Experiment}
\subsection{Datasets and Implementation Details}
\subsubsection{Datasets}
We evaluate our paradigm on both mathematical and coding tasks. 

\paragraph{Math Dataset}
For mathematical reasoning, we consider two representative benchmarks: the competition-level \textbf{AIME2025}~\cite{opencompass2025AIME} and the elementary-level \textbf{GSM8K} dataset~\cite{gsm8k}. Specifically, we sample 20 problems from AIME2025 and 10 problems from GSM8K to assess model performance across varying levels of difficulty.

\paragraph{Code Dataset}
For code generation and reasoning, we construct our evaluation set by sampling 20 problems of \textit{hard} difficulty from the \textbf{LeetCode} platform~\cite{leetcode}.

\subsubsection{Implementation Details}
\paragraph{Overall setup.}
We adopt a two-stage inference pipeline that delegates \texttt{think} to a non-autoregressive (NAR) diffusion language model and \texttt{answer} to either the same NAR model or a strong autoregressive (AR) model. Concretely, we use Mercury~\cite{mercuryCoder2025} as the NAR reasoner to produce compact, explicit reasoning traces, and then instantiate two routing variants:
(i) \textsc{NAR$\rightarrow$NAR} (\emph{Mercury-follow}): Mercury consumes its own \texttt{think} to produce the final \texttt{answer}; 
(ii) \textsc{NAR$\rightarrow$AR} (\emph{GPT5-follow}): GPT-5 (auto mode) consumes Mercury's \texttt{think} and emits the \texttt{answer}. 
Unless stated otherwise, we generate exactly one \texttt{think} trace per instance (no self-consistency sampling) to isolate the effect of division-of-labor.

%

\subsection{Experimental Results}

\input{latex/tables/main}

We evaluate on three sets: AIME2025 (20 items), GSM8K (10 items), and LeetCode-Hard (20 items). The results are shown in Table~\ref{tab:main_result}.
The pipeline first uses Mercury (NAR) to produce a compact think trace, then generates the final answer either with Mercury itself (NAR$\rightarrow$NAR) or with GPT-5 in auto mode (NAR$\rightarrow$AR). Scores are pass@1 success rates (\%) and the overall number is the unweighted average across the three sets.

Overall, routing the final answer to GPT-5 given a Mercury-generated plan yields substantial improvements on both math and code tasks. The largest gains appear on competition-level math, while easier arithmetic-compositional problems approach a ceiling but still benefit from the NAR plan. Code tasks also see clear advantages, indicating that a compact, globally coherent plan helps the AR decoder realize precise outputs.

\begin{enumerate}
    \item Across-the-board improvements: +40 on AIME2025 (50 vs.\ 10), +10 on GSM8K (100 vs.\ 90), +20 on LeetCode (95 vs.\ 75), with a +26 average lift (78 vs.\ 52).
    \item Strong advantage in harder regimes (AIME2025), suggesting the NAR think stage mitigates long-horizon derivation errors before AR realization.
    \item Near-ceiling GSM8K still benefits from the NAR plan, indicating complementary strengths even on simpler compositions.
    \item Code generation improves notably, supporting the division-of-labor design where NAR provides structured plans and AR ensures faithful surface realization.
\end{enumerate}

%% file: latex/tables/main.tex
\begin{table*}[t!]
\centering
\begin{tabular}{l|cc|c|c}
\toprule 
\multirow{2}{*}{\textbf{Method}} 
    & \multicolumn{2}{c|}{\textbf{Math}} 
    & \multicolumn{1}{c|}{\textbf{Code}} 
    & \multirow{2}{*}{\textbf{Average}} \\
    & AIME2025 & GSM8K & Leetcode & \\
\midrule
\multicolumn{5}{c}{\textit{NAR\&AR Framework (Our Paradigm)}}  \\
\midrule
GPT5+Mercury       & \textbf{50\%} & \textbf{100\%} & \textbf{95\%} & \textbf{78\%} \\
\midrule
\multicolumn{5}{c}{\textit{AR\&AR Framework (Baseline Paradigm)}}  \\
\midrule
Mercury+Mercury    & 10\% & 90\% & 75\% &52\% \\
\bottomrule
\end{tabular}
\caption{Main results on reasoning tasks like code and math. All numbers are success rates (\%). 
The \textbf{bold} entries denote the best results.
Our proposed NAR+AR paradigm (\textit{think with Mercury, answer with GPT-5}) achieves consistent improvements over the AR+AR baseline.}
\label{tab:main_result}
\vspace{-0.5cm}
\end{table*}